\documentclass[10pt,twocolumn,letterpaper]{article}

\usepackage{cvpr}              

\usepackage{graphicx}
\usepackage{amsmath}
\usepackage{amssymb}
\usepackage{booktabs}

\usepackage[pagebackref,breaklinks,colorlinks]{hyperref}

\usepackage[capitalize]{cleveref}
\crefname{section}{Sec.}{Secs.}
\Crefname{section}{Section}{Sections}
\Crefname{table}{Table}{Tables}
\crefname{table}{Tab.}{Tabs.}

\def\cvprPaperID{6} 
\def\confName{CVPR}
\def\confYear{2022}

\begin{document}
\title{Discriminability-enforcing loss to improve representation learning\vspace{-0.2cm}} 
%

\author{Florinel-Alin Croitoru \and Diana-Nicoleta Grigore \and Radu Tudor Ionescu\vspace{-0.2cm}\\
\and
University of Bucharest, Romania
}

%
%
%

%
\maketitle
\begin{abstract}
\vspace{-0.2cm}
During the training process, deep neural networks implicitly learn to represent the input data samples through a hierarchy of features, where the size of the hierarchy is determined by the number of layers. In this paper, we focus on enforcing the discriminative power of the high-level representations, that are typically learned by the deeper layers (closer to the output). To this end, we introduce a new loss term inspired by the Gini impurity, which is aimed at minimizing the entropy (increasing the discriminative power) of individual high-level features with respect to the class labels. Although our Gini loss induces highly-discriminative features, it does not ensure that the distribution of the high-level features matches the distribution of the classes. As such, we introduce another loss term to minimize the Kullback–Leibler divergence between the two distributions. We conduct experiments on two image classification data sets (CIFAR-100 and Caltech 101), considering multiple neural architectures ranging from convolutional networks (ResNet-17, ResNet-18, ResNet-50) to transformers (CvT). Our empirical results show that integrating our novel loss terms into the training objective consistently outperforms the models trained with cross-entropy alone, without increasing the inference time at all.
\vspace{-0.2cm}
\end{abstract}
%
%

\setlength{\abovedisplayskip}{4.0pt}
\setlength{\belowdisplayskip}{4.0pt}

\section{Introduction}
\label{sec:intro}
\vspace{-0.1cm}

Learning good data representations is a crucial point towards improving the performance of deep neural networks. Therefore, discovering new methods \cite{Caron-NIPS-2020, Chen-ICML-2020, Oord-CoRR-2018, Mesnil-ICML-2012, Goodfellow-NIPS-2011, Raffel-JMLR-2020,Huang-FN-2021, Khosla-NIPS-2020, Wen-ECCV-2016} to improve this capability is an important goal for the research community. \emph{Representation learning} \cite{Bengio-TPAMI-2013, Goodfellow-DLMIT-2016} is a vast domain studying this direction, and it covers multiple topics ranging from the design of neural architectures \cite{Hinton-SC-2006,Kingma-ICLR-2014, Higgins-ICLR-2017,Goodfellow-NIPS-2014} to the development of learning paradigms \cite{Caron-NIPS-2020, Chen-ICML-2020,Oord-CoRR-2018,Samuli-ICLR-2017, Tarvainen-NIPS-2017}. 

Our study fits in this domain, as our focus is on learning better high-level representations by enforcing their discriminability towards the target classes, in the context of supervised learning.

During the training process, deep neural networks implicitly learn to represent the input data samples through a hierarchy of features. As shown in \cite{Zeiler-ECCV-2014}, the features closer to the input tend to encode low-level information, e.g.~edges, corners, stains, while those closer to the output tend to encode high-level information, e.g.~object parts or even entire objects. While the low-level features are generic to all object classes, the high-level features should be specialized in discriminating between object classes. To further boost the discriminative power of the high-level representations, we propose to add two new loss terms to the objective function. The first loss term relies on the Gini impurity to minimize the entropy of individual high-level features with respect to the class labels. The entropy of the features (activation maps) gets minimized as they become more discriminative towards certain classes. Although our Gini loss induces highly-discriminative features, it does not ensure that the distribution of the high-level features matches the distribution of the classes. In other words, it might lead to a disproportionate number of features being specialized on a certain class. As such, we introduce another loss term to minimize the Kullback–Leibler (KL) divergence between the feature and the class distributions.

We conduct experiments on two image classification data sets (CIFAR-100 and Caltech 101), considering multiple neural architectures ranging from convolutional networks (ResNet-17, ResNet-18, ResNet-50) \cite{He-CVPR-2016} to transformers (CvT) \cite{Wu-ICCV-2021}. Our empirical results show that integrating our novel loss terms into the training objective consistently outperforms the models trained with cross-entropy alone. The accuracy gains come at no computational cost during inference, making the use of deeper and more computationally intensive models unnecessary.

\noindent {\bf Contribution.}
In summary, our contribution is threefold:
\begin{itemize}
\item \vspace{-0.2cm} We introduce a new loss based on the Gini impurity, which boosts the discriminative power of high-level features.
\item \vspace{-0.2cm} We introduce a new loss based on the KL divergence, which ensures that the distribution of the high-level representation matches the class distribution.
\item \vspace{-0.2cm} We present empirical evidence showing the benefits of our approach across neural architectures and data sets.
\end{itemize}

\vspace{-0.25cm}
\section{Related work}
\label{sec:relatedart}
\vspace{-0.15cm}

Representation learning is a vast research topic \cite{Bengio-TPAMI-2013, Goodfellow-DLMIT-2016} and, through deep learning methods, it has attained significant progress in various domains, such as computer vision \cite{He-CVPR-2016, Wu-ICCV-2021} and natural language processing \cite{Devlin-NAACL-2019, Radford-GPT-2018}. Aside from these achievements, learning good representations is important because they unlock solutions for other scenarios via transfer learning \cite{Mesnil-ICML-2012, Goodfellow-NIPS-2011, Raffel-JMLR-2020}, domain adaptation \cite{Glorot-ICML-2011, Chen-2012-ICML} or modality learning \cite{Srivastava-JMLR-2014, Hao-EMNLP-2019}. What constitutes a good representation is described in \cite{Bengio-TPAMI-2013, Goodfellow-DLMIT-2016} through a set of properties.

A considerable effort in learning better representations has been made in unsupervised settings, where architectures such as auto-encoders \cite{Hinton-SC-2006}, variational auto-encoders (VAE) \cite{Kingma-ICLR-2014, Higgins-ICLR-2017}, and generative adversarial networks (GANs) \cite{Goodfellow-NIPS-2014, Chen-NIPS-2016, Radford-ICLR-2016} have been proposed. Deep convolutional GAN (DCGAN) \cite{Radford-ICLR-2016}, InfoGAN \cite{Chen-NIPS-2016} and $\beta-$VAE \cite{Higgins-ICLR-2017} are able to learn expressive representations with multiple explanatory factors \cite{Bengio-TPAMI-2013}, such as emotion and hairstyle in images of human faces, or digit type and rotation in MNIST images. However, despite these improvements, in computer vision, the state-of-the-art solutions in image recognition \cite{ He-CVPR-2016,Wu-ICCV-2021} still employ supervised transfer learning to learn better features. Semi-supervised learning \cite{Samuli-ICLR-2017, Tarvainen-NIPS-2017} utilizes both labeled and unlabeled data for learning representations. Other approaches, such as pre-trained language models \cite{Devlin-NAACL-2019,Radford-GPT-2018} and contrastive learning frameworks \cite{Caron-NIPS-2020, Chen-ICML-2020, Oord-CoRR-2018}, fall under the self-supervised paradigm. Interestingly, the method presented in \cite{Caron-NIPS-2020} is very close in terms of image classification performance to the supervised counterpart. In natural language processing, a common technique is to pre-train language models on an extensive unsupervised corpus to learn language representations, and then fine-tune them on supervised tasks. Popular examples such as BERT \cite{Devlin-NAACL-2019} and GPT \cite{Radford-GPT-2018} brought notable improvements on multiple language understanding tasks.

Closer to our work, there are several methods that use other objective functions besides cross-entropy to impose certain properties on the learned features \cite{Huang-FN-2021, Khosla-NIPS-2020, Wen-ECCV-2016}. In \cite{Khosla-NIPS-2020}, the contrastive loss from the self-supervised case \cite{Caron-NIPS-2020} is adapted to supervised scenarios, the inflicted properties being smoothness and space coherence \cite{Bengio-TPAMI-2013, Goodfellow-DLMIT-2016}. The same can be said about other works \cite{Huang-FN-2021, Wen-ECCV-2016}, where the additional center loss penalizes a large variation within each class. Our loss function is distinct from related methods by not being directly connected to the distances or similarities in the representation space. Instead, it simply enforces the high-level features to be discriminative across classes. To the best of our knowledge, we are the first to employ a discriminability-enforcing loss to learn better high-level features.

\vspace{-0.25cm}
\section{Method}
\label{sec:method}
\vspace{-0.15cm}

The cross-entropy loss is broadly employed as an objective function for training deep classifiers, being able to push any model to learn separable representations across classes. However, the training objective is not necessarily the most suitable for generalizing to unseen data, because the implicitly induced separability can be fragile. We address this shortage by optimizing an auxiliary loss function that can compel any model to learn more discriminative representations, meaning more class-oriented representations. Further in this section, we describe our approach comprising two novel components added to the loss function, while also explaining their roles.

The Gini impurity is often used to estimate the discriminative power of features in random forests, when a feature needs to be selected for a certain node while building a decision tree. Inspired by this choice, we propose to introduce a loss based on the Gini impurity to enforce the discriminative power of a set of convolutional activation maps $A \in \mathbb{R}^{m\times c \times h \times w}$, where $m$ is the number of training data samples, $c$ is the number of channels, and $h$ and $w$ are the height and width of an activation map. Our Gini loss is defined as follows: 
\begin{equation}
\mathcal{L_{\scriptsize{\mbox{Gini}}}} = \frac{1}{c} \sum_{i=1}^c \left( 1 - \sum_{j=1}^n \bar{s}_{ij}^{\:2} \right),
\end{equation}
where $n$ represents the number of classes, and $\bar{s}_{ij}$ are elements of the matrix $\bar{S}$ having $c \times n$ components, whose components are probabilities computed as follows:
\begin{equation}
\bar{s}_{ij} =\frac{s_{ij}}{\sum_{k=1}^n s_{ik}}, \forall i \in \{1,2,...,c\}, j \in \{1,2,...,n\}.
\end{equation}
Here, $s_{ij}$ represents the average activation on channel $i$ for class $j$. The average activations $s_{ij}$ are elements of the matrix $S$ of size $c \times n$, which is computed as follows:
\begin{equation}
S = M^{\top} \cdot \bar{Y},
\end{equation}
where $M$ is a matrix of $m \times c$ components resulted after applying global average pooling over the set of activation maps $A$, and $\bar{Y}$ is a matrix of $m \times n$ components denoted as $\bar{y}_{ij}$, which are computed as follows:
\begin{equation}
\bar{y}_{ij} = \frac{y_{ij}}{\sum_{k=1}^m y_{kj}}, \forall i \in \{ 1,2,...,m \}, j \in \{1,2,...,n\},
\end{equation}
where $y_{ij}$ are elements of the matrix $Y$ of size $m \times n$ containing the target labels (as $n$-dimensional one-hot encodings) for the $m$ training samples.

While constraining the activation maps to be more discriminative towards certain classes, $\mathcal{L_{\scriptsize{\mbox{Gini}}}}$ does not influence the distribution of the activation maps over classes. Hence, we might end up having many discriminative activation maps for some classes and none for other classes. To ensure that the distribution of discriminative activation maps matches the training class distribution, we propose to introduce a loss based on the Kullback–Leibler divergence, defined as: 
\begin{equation}
\mathcal{L_{\scriptsize{\mbox{KL}}}} = \mbox{KL} \left( H^S, H^Y \right),
\end{equation}
where $H^S$ and $H^Y$ are histograms of $n$ bins representing the discrete distributions of the activations maps and the class labels, respectively. Each component $h_i^S \in H^S$ is computed by summing up and normalizing the average activations in $\bar{S}$, as follows:
\begin{equation}
h_i^S = \frac{\sum_{j=1}^c \bar{s}_{ji}}{\sum_{k=1}^n \sum_{j=1}^c \bar{s}_{jk}}, \forall i \in \{1,2,...,n\}.
\end{equation}
Similarly, each component $h_i^Y \in H^Y$ is computed by summing up and normalizing the one-hot labels in $Y$, as follows:
\begin{equation}\label{eq_hy}
h_i^Y = \frac{\sum_{j=1}^m y_{ji}}{m}, \forall i \in \{1,2,...,n\}.
\end{equation}
In Eq.~\eqref{eq_hy}, we use the fact that the sum of all components in $Y$ is equal to $m$ (the number of training samples).

When integrating our approach into a neural model having its own loss function $\mathcal{L}_*$, our losses can simply be added to the respective loss, resulting in a new loss function that comprises all terms:
\begin{equation}\label{eq_loss_total}
\mathcal{L}_{\mbox{\scriptsize{total}}} = \mathcal{L}_* + \lambda_1 \cdot \mathcal{L}_{\mbox{\scriptsize{Gini}}} + \lambda_2 \cdot \mathcal{L}_{\mbox{\scriptsize{KL}}},
\end{equation}
where $\lambda_1, \lambda_2  \in \mathbb{R}^+$ are hyperparameters that control the importance of each loss term with respect to the original loss $\mathcal{L}_*$. For simplicity, we set $\lambda_1=\lambda_2$ in our experiments.

\vspace{-0.1cm}
\section{Experiments}
\label{sec_experiments}
\vspace{-0.05cm}

\subsection{Data Sets}

\noindent {\bf CIFAR-100.} The CIFAR-100 data set \cite{Krizhevsky-TECHREP-2009}  consists of 60,000 color images of 32$\times$32 pixels. The images are grouped into 20 superclasses that are further divided into 100 mutually exclusive classes, each containing 500 training images and 100 test images. We kept a subset of 4,000 images (40 images per category) from the training set for validation. 

\noindent {\bf Caltech 101.} The Caltech 101 data set \cite{Fei-CVPRW-2004} contains a total of 9,146 pictures of objects from 101 different categories and a special background class. Each category includes between 40 and 800 images, with the more common or important classes being better represented. The images have varying sizes of around 300$\times$200 pixels. 

\vspace{-0.1cm}
\subsection{Evaluation Setup} 
\vspace{-0.05cm}

\noindent {\bf In-domain and cross-domain evaluation.} We conduct both in-domain and cross-domain experiments to exhibit the generalization capacity of our approach across different scenarios. The in-domain experiments are conducted on CIFAR-100. To perform cross-database assessments of the neural models, we create a custom subset of Caltech 101, which contains all the categories from the intersection with CIFAR-100. Due to the imperfect automatic mapping between category names, we performed manual label matching. Finally, the resulting Caltech 101 subset consists of 12 object categories. 
For the cross-domain experiments, the models are trained on CIFAR-100 and evaluated on the Caltech 101 subset.

\noindent {\bf Evaluation measure.} We quantify the performance of neural models in terms of the classification accuracy. We train and evaluate each neural network in $5$ trials, reporting its average accuracy and standard deviation.

\noindent {\bf Baselines.}
We study the impact of our method on two types of deep architectures, namely residual neural networks (ResNets) \cite{He-CVPR-2016} and Convolutional vision Transformers (CvT) \cite{Wu-ICCV-2021}. ResNets \cite{He-CVPR-2016} form a class of convolutional neural networks approaching the gradient propagation problem, which appears in very deep models, with residual connections implemented as identity or projection mappings. In this work, we employ three such models that vary in depth: ResNet-17, ResNet-18 and ResNet-50. 
We derive ResNet-17 from ResNet-18 by removing the fully connected layer at the end and adjusting the last convolutional layer to have the number of filters equal to the number of class labels, thus obtaining a lighter architecture. 

CvT \cite{Wu-ICCV-2021} is a transformer-based architecture that organizes the transformer blocks into stages, incorporating the convolution operation via two methods, called convolutional token embedding and convolutional projection. 
In our experiments, we employ the lighter CvT-7 architecture, which contains three stages. The first stage has one transformer block, the second has two blocks and the third has four blocks. In each stage, the convolutional token embedding layers have $32$, $128$ and $258$ filters, respectively. The number of heads of the Multi-Head Self-Attention module also varies between stages: the blocks in the first stage have one head, those in the second stage have two heads, and those in the third stage have six heads.

 
\noindent {\bf Hyperparameter tuning.} We trained both ResNet-17 and ResNet-18 for $200$ epochs on mini-batches of $500$ examples. We started with a learning rate of $10^{-3}$ and periodically reduced it by a factor of $0.5$ after $10$ epochs of no improvement, with a threshold of $10^{-2}$. We used Adam to optimize both models. In a similar fashion, we trained ResNet-50 for $250$ epochs on mini-batches of size $200$. We optimized the model using stochastic gradient descent with momentum, setting the momentum rate to $0.9$. We started with a learning rate of $10^{-1}$ and a weight decay of $5 \cdot 10^{-4}$. At epochs $60$, $120$, $160$, and $200$, we reduced the learning rate by a factor of $0.2$. For CvT, we set the number of epochs to $200$ and the mini-batch size to $200$. We optimized CvT using AdaMax, with an initial learning rate of $2 \cdot 10^{-3}$. All baselines are trained using the categorical cross-entropy as the loss function $\mathcal{L}_*$.

\begin{figure*}[!t]
\begin{center}
\centering
\includegraphics[width=1.0\linewidth]{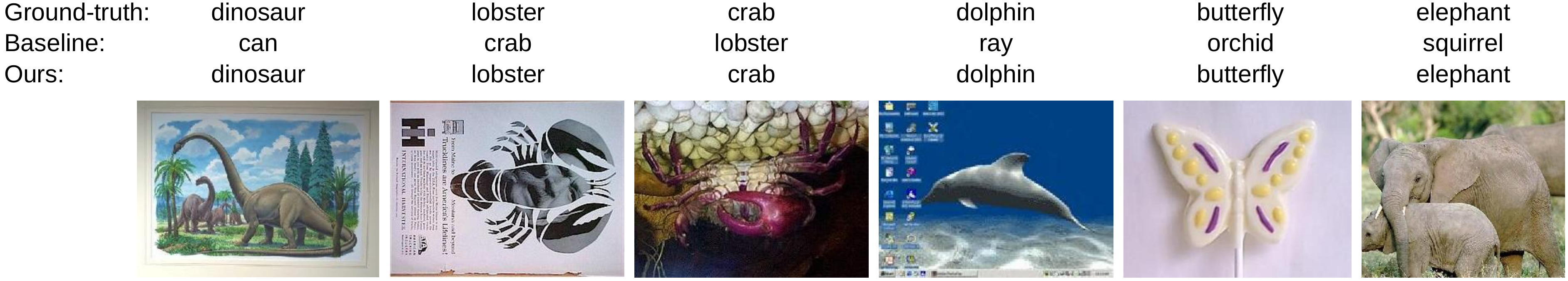}
\vspace{-0.65cm}
\caption{Examples from Caltech 101 that are wrongly classified by a ResNet-50 trained on CIFAR-100 using cross-entropy (baseline). The model is able to correctly label the samples upon introducing our novel loss. Best viewed in color.}
\label{fig_pipeline}
\end{center}
\vspace{-0.5cm}
\end{figure*}

Aside from setting the common hyperparameters mentioned above, our method requires setting additional values, such as the two weights that control the importance of our new loss terms. In a preliminary set of validation experiments, we observed that setting both $\lambda_1$ and $\lambda_2$ to $0.5$ works reasonably well across different neural architectures. Another necessary configuration is the epoch index from which we begin to introduce $\mathcal{L_{\scriptsize{\mbox{Gini}}}}$ and $\mathcal{L_{\scriptsize{\mbox{KL}}}}$. We set the starting epoch to $10$ in all our experiments. Our loss terms are applied on the last convolutional layer of each residual architecture. Analogously, for CvT, we employ them on the convolutional token embedding layer from the last stage.

\begin{table}[!t]
\setlength\tabcolsep{4.2pt}
\caption{Accuracy scores (in \%) on CIFAR-100 and Caltech 101. The baselines trained with cross-entropy alone are compared with our models trained with the loss defined in Eq.~\eqref{eq_loss_total}. Top scores for each architecture are highlighted in bold.\vspace{-0.5cm}}
\label{tab_Results}
\begin{center}
\begin{tabular}{|l|c|c|}
\hline
{Model}                     & {CIFAR-100}               & {Caltech 101} \\ 
\hline
\hline
ResNet-17 (baseline)        & $71.88\pm 0.36$	        & $63.22\pm 0.68$\\
ResNet-17 (ours)            & $\mathbf{73.00}\pm 0.20$  & $\mathbf{63.90}\pm 1.56$	 \\
\hline
ResNet-18 (baseline)        & $71.47\pm 0.36$           & $63.04\pm 1.35$\\
ResNet-18 (ours)            & $\mathbf{72.34}\pm 0.13$  & $\mathbf{64.33}\pm 0.84$\\
\hline
ResNet-50 (baseline)        & $76.45\pm 0.51$           & $70.14\pm 1.26$\\
ResNet-50 (ours)            & $\mathbf{76.92}\pm 0.61$  & $\mathbf{70.68}\pm 0.74$ \\
\hline
CvT-7 (baseline)            & $63.34\pm 0.56$           & $53.35\pm 1.26$\\  
CvT-7 (ours)                & $\mathbf{63.76}\pm 0.25$  & $\mathbf{53.79}\pm 0.89$ \\ 
\hline
\end{tabular}
\end{center}
\vspace{-0.5cm}
\end{table}

\vspace{-0.1cm}
\subsection{Results}
\vspace{-0.05cm}

We present the results on CIFAR-100 and Caltech 101 in Table~\ref{tab_Results}. First, we observe that all the models trained with our new loss perform better than their counterparts based solely on cross-entropy. In addition, our models yield better accuracy scores in the cross-domain setup, demonstrating their capability of learning more general representations than the baselines. Another remark is that the standard deviation is generally lower for our approach, suggesting that the results are more stable across multiple runs. We underline that the accuracy gains come at no additional computational cost during inference. Moreover, we observe that the lighter ResNet-17 architecture usually attains better performance than ResNet-18, indicating that more efficient models can also be more effective.

In Table~\ref{tab_comparison}, we present ablation results that show the effect of each loss term on the final accuracy. The ablation study indicates that using our loss terms independently leads to performance drops. The results demonstrate the importance of jointly using our novel loss terms, confirming the necessity of using both terms to achieve the desired accuracy gains.

We illustrate a few qualitative results in Figure~\ref{fig_pipeline}. The most noteworthy examples are the second and third images (counting from left to right), which show that the baseline confuses two similar classes, lobster and crab, while our model is able to correctly tag the respective examples.

\section{Conclusion}
\label{sec:conclusion}

In this paper, we proposed a novel approach to boost the discriminative power of high-level representations in convolutional and transformer architectures. Our approach is based on jointly integrating two novel loss terms, one that enforces the discriminability of individual features and another that ensures the alignment between the high-level feature distribution and the class distribution. We presented empirical evidence indicating that our approach brings accuracy gains for multiple neural architectures across different evaluation scenarios.

In future work, we aim to evaluate our approach on further benchmarks and neural architectures. We also aim to extend our approach to regression tasks, which could require modifying our loss terms.

\begin{table}[!t]
\setlength\tabcolsep{4.2pt}
\caption{Ablation results for ResNet-18 on CIFAR-100, obtained by excluding various loss terms from Eq.~\eqref{eq_loss_total}. $\mathcal{L}_*$ represents the categorical cross-entropy.\vspace{-0.5cm}}
\label{tab_comparison}
\begin{center}
\begin{tabular}{|c|c|c|c|}
\hline
\bf $\mathcal{L}_*$ & $\mathcal{L}_{\mbox{\scriptsize{Gini}}}$ & $\mathcal{L}_{\mbox{\scriptsize{KL}}}$ & Accuracy\\
\hline
\hline
\checkmark &            &           & $71.47\pm 0.36$\\
\hline
\checkmark & \checkmark &           & $69.87\pm 0.50$\\
\hline
\checkmark &            & \checkmark & $69.72\pm 0.38$	\\
\hline
\checkmark & \checkmark & \checkmark & $72.34\pm 0.13$\\
\hline
\end{tabular}
\end{center}
\vspace{-0.5cm}
\end{table}

\vspace{-0.15cm}
\section*{Acknowledgments}
\vspace{-0.15cm}
The research leading to these results has received funding from the NO Grants 2014-2021, under project ELO-Hyp contract no. 24/2020. This article has also benefited from the support of the Romanian Young Academy, which is funded by Stiftung Mercator and the Alexander von Humboldt Foundation for the period 2020-2022.

\vspace{-0.15cm}


{\small
\bibliographystyle{ieee_fullname}
\bibliography{references}
}
\end{document}